\documentclass[lettersize,journal]{IEEEtran}
\usepackage{amsmath,amsfonts}
\usepackage{algorithmic}
\usepackage{algorithm}
\usepackage{array}
\usepackage[caption=false,font=normalsize,labelfont=sf,textfont=sf]{subfig}
\usepackage{textcomp}
\usepackage{stfloats}
\usepackage{url}
\usepackage{verbatim}
\usepackage{graphicx}
\usepackage{cite}

\usepackage{booktabs}       
\usepackage{xcolor}         
\usepackage{nicefrac}       
\usepackage{microtype}      

\usepackage{amssymb}
\usepackage{listings}
\usepackage{setspace}

\usepackage{rotating}
\usepackage{enumitem}

\usepackage{tabularx}
\usepackage{adjustbox}

\graphicspath{ {./ffe_images/} }

\newcommand{\DD}[2]{\frac{\mathrm{d} #1}{\mathrm{d} #2}}
\newcommand{\DDT}[1]{\DD{#1}{t}}

\begin{document}

\title{Signal Propagation: A Framework for Learning and Inference In a Forward Pass}

\author{
    Adam Kohan,
    Edward A. Rietman,
    Hava T. Siegelmann
    \thanks{Adam Kohan, Edward Rietman, and Hava Siegelmann are with the Biologically Inspired Neural and Dynamical Systems Laboratory, College of Information and Computer Sciences, University of Massachusetts Amherst (e-mail: akohan@cs.umass.edu, erietman@cs.umass.edu, hava@cs.umass.edu)}
}

\markboth{IEEE TRANSACTIONS ON NEURAL NETWORKS AND LEARNING SYSTEMS}%
{Kohan \MakeLowercase{\textit{et al.}}: Signal Propagation Learning}


\maketitle

\begin{abstract}
    We propose a new learning framework, signal propagation (sigprop), for propagating a learning signal and updating neural network parameters via a forward pass, as an alternative to backpropagation. In sigprop, there is only the forward path for inference and learning. So, there are no structural or computational constraints necessary for learning to take place, beyond the inference model itself, such as feedback connectivity, weight transport, or a backward pass, which exist under backpropagation based approaches. That is, sigprop enables global supervised learning with only a forward path. This is ideal for parallel training of layers or modules. In biology, this explains how neurons without feedback connections can still receive a global learning signal. In hardware, this provides an approach for global supervised learning without backward connectivity. Sigprop by construction has compatibility with models of learning in the brain and in hardware than backpropagation, including alternative approaches relaxing learning constraints. We also demonstrate that sigprop is more efficient in time and memory than they are. To further explain the behavior of sigprop, we provide evidence that sigprop provides useful learning signals in context to backpropagation. To further support relevance to biological and hardware learning, we use sigprop to train continuous time neural networks with Hebbian updates, and train spiking neural networks with only the voltage or with biologically and hardware compatible surrogate functions.
\end{abstract}


\section{Introduction}
\IEEEPARstart{T}{he} success of deep learning is attributed to the backpropagation of errors algorithm \cite{rumelhart1986learning} for training artificial neural networks. However, the constraints necessary for backpropagation to take place are incompatible with learning in the brain and in hardware, are computationally inefficient for memory and time, and bottleneck parallel learning. These learning constraints under backpropagation come from calculating the contribution of each neuron to the network's output error. This calculation during training occurs in two phases. First, the input is fed completely through the network storing the activations of neurons for the next phase and producing an output; this phase is known as the forward pass. Second, the error between the input's target and network's output is fed in reverse order of the forward pass through the network to compute parameter updates using the stored neuron activations; this phase is known as the backward pass.

These two phases of learning have the following learning constraints. The forward pass stores the activation of every neuron for the backward pass, increasing memory overhead. The forward and backward passes need to complete before receiving the next inputs, thereby pausing resources. Network learning parameters can only be updated after and in reverse order of the forward pass, which is sequential and synchronous. The backward pass requires its own feedback connectivity to every neuron, increasing structural complexity. The feedback connectivity needs to have weight symmetry with forward connectivity, known as the weight transport problem. The backward pass uses a different type of computation than the forward pass, adding computational complexity. In total, these constraints prohibit parallelization of computations during learning, increase memory usage, run time, and the number of computations, and bound the network structure.

These learning constraints under backpropagation are difficult to reconcile with learning in the brain\cite{marblestone2016toward,grossberg1987competitive}. Particularly, the backward pass is considered to be problematic \cite{marblestone2016toward,grossberg1987competitive,crick1989recent,lee2016training,scellier2017equilibrium} as (1) the brain does not have the comprehensive feedback connectivity necessary for every neuron (2) neither is neural feedback known to be a distinct type of computation, separate from feedforward activity and (3) the feedback and feedforward connectivity would need to have weight symmetry.

These learning constraints also hinder efficient implementations of backpropagation and error based learning algorithms on hardware \cite{neftci2017event,bouvier2019spiking}: (1) weight symmetry is incompatible with elementary computing units which are not bidirectional, (2) the transportation of non local weight and error information requires special communication channels in hardware, and (3) spiking equations are non-derivable, non-continuous. Hardware implementations of learning algorithms may provide insight into learning in the brain. An efficient, empirically competitive algorithm to backpropagation on hardware will likely parallel learning in the brain.

All of these constraints can be categorized by their overall effect on learning for a network as follows. (a) Backwardpass unlocking would allow for all parameters to be updated in parallel after the forward pass has completed. (b) Forwardpass unlocking would allow for individual parameters to be asynchronously updated once the forward pass has reached them, without waiting for the forward pass to complete. These categories directly reference parallel computation, but also have implications on network structure, memory, and run-time. For example, backwardpass locking implies top-down feedback connectivity. Similar terminology was used in \cite{jaderberg2017decoupled}, where (a) is backward locking and (b) is update locking. Alternative learning approaches to address backwardpass and forwardpass unlocking have been proposed, refer to Section \ref{sec:relax} and Figure \ref{fig:models}, but do not solve all of these constraints and are based on relaxing learning constraints under backpropagation.

We propose a new learning framework, signal propagation (SP or sigprop), for propagating a learning signal and updating neural network parameters via a forward pass. Sigprop has no constraints on learning, beyond the inference model itself, and is completely forwardpass unlocked. At its core, sigprop generates targets from learning signals and then re-uses the forward path to propagate those targets to hidden layers and update parameters. Sigprop has the following desirable features. First, inputs and learning signals use the same forward path, so there are no additional structural or computational requirements for learning, such as feedback connectivity, weight transport, or a backward pass. Second, without a backward pass, the network parameters are updated as soon as they are reached by a forward pass containing the learning signal. Sigprop does not block the next input or store activations. So, sigprop is ideal for parallel training of layers or modules. Third, since the same forwardpass used for inputs is used for updating parameters, there is only one type of computation. Compared with alternative approaches, sigprop addresses all of the above constraints, and does so with a global learning signal.

Our work suggests that learning signals can be fed through the forward path to train neurons. Feedback connectivity is not necessary for learning. In biology, this means that neurons who do not have feedback connections can still receive a global learning signal. In hardware, this means that global learning (e.g supervised or reinforcement) is possible even though there is no backward connectivity.

This paper is organized as follows. In Section \ref{sec:relax}, we detail the improvements on relaxing learning constraints of sigprop over alternative approaches. In Section \ref{sec:ffl}, we introduce the signal propagation framework and learning algorithm. In Section \ref{sec:exp}, we describe experiments evaluating the accuracy, run time, and memory usage of sigprop. We also demonstrate that sigprop can be trained with a sparse learning signal. In Section \ref{sec:contime}, we demonstrate that sigprop provides a useful learning signal that becomes increasingly similar to backpropagation as training progresses. We also demonstrate that sigprop can train continuous time neural networks, and with a Hebbian plasticity mechanism to update parameters in hidden layers, as further support of its relevance to biological learning. In Section \ref{sec:spike}, we demonstrate that sigprop directly trains Spiking Neural Networks, with or without surrogate functions, as further support of its relevance to hardware learning.

\section{Relaxing Constraints on Learning} \label{sec:relax}
\begin{figure*}[!t]
\centering
\includegraphics[scale=.40,trim = 0 10 42 0,clip]{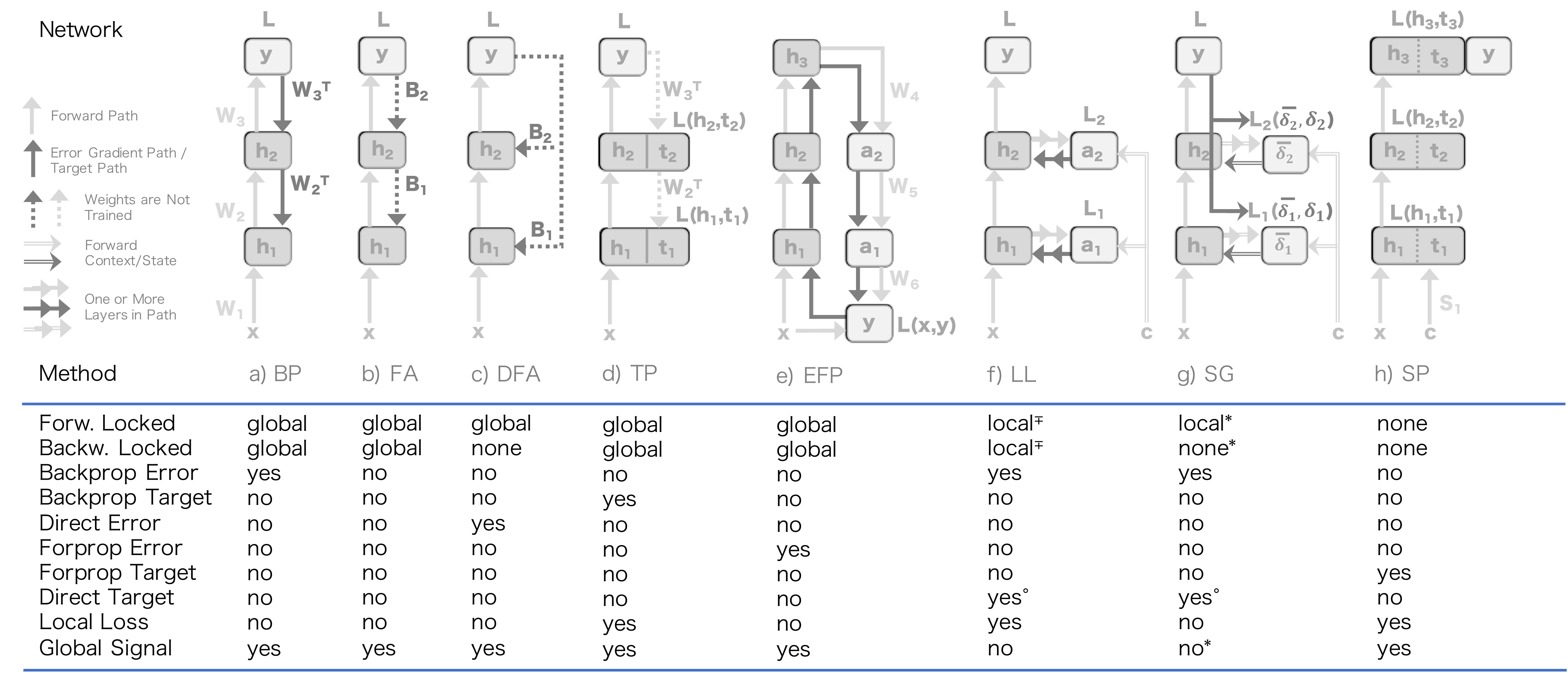}
\caption{Comparison of Learning Algorithms Relaxing Learning Constraints Under Backpropagation. \textbf{a)} the backpropagation algorithm \textbf{b,c)} the feedback alignment and direct feedback alignment algorithms. FA based algorithms do not solve forwardpass locking and require additional connectivity. \textbf{d)} target propagation uses a single type of computation for training and inference, but is forwardpass locked and requires feedback connectivity. \textbf{e)} error forward propagation for closed loop systems or autoencoders reuses the forward connectivity to propagate error, but is otherwise similarly constrained as backpropagation. \textbf{f)} local learning with layer-wise training using auxiliary classifiers. $^{\mp}$LL is forwardpass and backwardpass locked at the layer level as the auxiliary networks use backpropagation. Backpropagation in the auxiliary networks may be substituted with an alternative model, such as FA. \textbf{g)} the synthetic gradients algorithm. *SG based algorithms are only forwardpass and backwardpass unlocked after learning to predict the synthetic gradient. \textbf{h)} the signal propagation learning algorithm presented in this work. SP feeds the learning signal forward through the network to solve the weight transport and forwardpass locking problems without requiring additional connectivity requirements. For SP, taking $t_3$ with $h_3$ produces $y$, however a classification layer may also be used Fig. \ref{fig:fflversions}. \textbf{Table)} Direct Error and Direct Target means that a model uses the error or target directly at layer $h_i$. $^{\circ}$Direct target can be substituted in LL and SG, with direct error or temporary use of backpropagation for example. Forprop stands for forward propagation. Forprop error and Forprop target means the model uses the error or target starting at the input layer, instead of starting at the output layer as is done in backpropagation. Global Signal means the learning signal is propagated through the network instead of sent directly to or formed at each hidden layer. \textbf{Networks)} The light grey arrows indicate the feed forward path. Dark grey arrows indicates error gradient or target paths. If the dark grey arrow pass through a layer, the weights are not trained by the error gradient or target. Dotted lines indicate the weights are not trained. Double lines, light or dark grey, are forwarding the context $c$ or state $h_i$, without modification. Double arrows indicate going through one or more intermediate hidden layers. $W_i$ and $S_i$ are trained weights and $B_i$ are fixed random weights. There are versions of these models where $B_i$ is trained to be the transpose of $W_i$. The loss function is $L$ and takes the output of the previous layer and possibly some target $y^*$ when unspecified. The target generator layer $S_1$ generates the initial training target $t_i$ from a learning signal, which is some privileged information or context $c$, usually the label in supervised learning. The gradient is $\delta$ and the synthetic gradient is $\hat{\delta}$. Auxiliary networks are represented by the double arrows going into $a_i$ and $\hat{\delta}_i$.}
    \label{fig:models}
\end{figure*}

Signal propagation (sigprop) is a new approach that imposes no learning constraints, beyond the inference model itself, while providing a global learning signal. Alternative approaches, in contrast, are based on relaxing the learning constraints under backpropagation. This is a view by which we can arrive at sigprop: once the learning constraints under backpropagation are done away with, the simplest explanation to provide a global learning is to use the forward path, the path constructing the inference model; that is, project the learning signal through the same path as the inputs. Here, we discuss alternative approaches, compare the variations of constraints they relax, and see the difference of removing constraints entirely, which results in the improvements shown under sigprop. Refer to Fig \ref{fig:models} for a visual comparison.

Feedback Alignment (FA), Fig \ref{fig:models}b uses fixed random weights to transport error gradient information back to hidden layers, instead of using symmetric weights \cite{lillicrap2016random}. It was shown that the sign concordance between the forward and feedback weights is enough to deliver effective error signals \cite{liao2016important,guerguiev2017towards,neftci2017event}. During learning, the forward weights move to align with the random feedback weights and have approximate symmetry, forming an angle below $90^\circ$. FA addresses the weight transport problem, but remains forwardpass and backwardpass locked. Direct Feedback Alignment (DFA), Fig \ref{fig:models}c propagates the error directly to each hidden layer and is additionally backwardpass unlocked \cite{nokland2016direct}. Sigprop improves on DFA and is forwardpass unlocked. DFA performs similarly to backpropagation on CIFAR-10 for small fully-connected networks with dropout, but performs more poorly for convolutional neural networks. Sigprop performs better than DFA and FA for convolutional neural networks.

FA based algorithms also rely on systematic feedback connections to layers and neurons. Though it is possible \cite{lillicrap2016random,scellier2017equilibrium,guerguiev2017towards}, there is no evidence in the neocortex of the comprehensive level of connectivity necessary for every neuron (or layer) to receive feedback (reciprocal connectivity). With sigprop, we introduce an algorithm capable of explaining how neurons without feedback connections learn. That is, neurons without feedback connectivity receive feedback through their feedforward connectivity.

An alternative approach that minimizes feedback connectivity is Local Learning (LL), Fig \ref{fig:models}f. In LL algorithms \cite{nokland2019training,belilovsky2020decoupled,kaiser2020synaptic}, layers are trained independently by calculating a separate loss for each layer using an auxiliary classifier per layer. LL algorithms have achieved performance close to backpropagation on CIFAR-10 and is making progress on ImageNet. It trains each layer and auxiliary classifier with backpropagation. At the layer level, it has the weight transport problem and is forwardpass and backwardpass locked. In \cite{nokland2019training}, FA is used to backwardpass unlock the layers. It does not use a global learning signal, but learns greedily. In another approach, Synthetic Gradients (SG), Fig \ref{fig:models}g are used to train layers independently \cite{jaderberg2017decoupled, czarnecki2017understanding}. SG algorithms train auxiliary networks to predict the gradient of the backward pass from the input, the synthetic gradient. Similar to LL, SG methods trains the auxiliary networks using backpropagation. Until the auxiliary networks are trained, it has the weight transport problem and is forwardpass and backwardpass locked at the network level. In contrast, sigprop is completely forwardpass unlocked, combines a global learning signal with local learning, is compatible with learning in hardware where there is no backward connectivity, and compatible with models of learning in the brain where comprehensive feedback connectivity is not seen, including projections of the targets to hidden layers.

Forwardpass unlocked algorithms do not necessarily address the limitations in biological and hardware learning models, as they have different types of computations for inference and learning. In sigprop, the approach to having a single type of computation for inference and learning is similar to target propagation. Target Propagation (TP), Fig \ref{fig:models}d \cite{lee2015difference, bengio2014auto} generates a target activation for each layer instead of gradients by propagating backward through the network. It requires reciprocal connectivity and is forwardpass and backwardpass locked. In contrast, sigprop generates a target activation at each layer by going forward through the network. An alternative approach, Equilibrium Propagation (EP) is an energy based model using a local contrastive Hebbian learning with the same computation in the inference and learning phases \cite{scellier2017equivalence,xie2003equivalence, scellier2017equilibrium}. The model is a continuous recurrent neural network that minimizes the difference between two fixed points: when receiving an input only and when receiving the target for error correction. EP is closer to a framework, wherein symmetric and random feedback (FA) weights work \cite{scellier2018extending}. These models of EP still require comprehensive connectivity for each layer and are forwardpass locked. We demonstrate that sigprop works in the EP framework without these problems, more closely modeling neural networks in the brain.

Another approach that reuses the forward connectivity for learning, as is we do in sigprop, is Error Forward Propagation, Fig \ref{fig:models}e \cite{hirasawa1996forward,williams1990gradient,ohama2004forward,ohama2005forward,heinz1995pipelined,ohama2017parallel}. Error forward propagation is for closed loop control systems or autoencoders. In either case, the output of the network is in the same space as the input of the network. These works calculate an error between the output and input of the network and then propagate the error forward through the network, instead of backward, calculating the gradient as in error backpropagation. Error forward propagation is backwardpass locked and forwardpass locked. It also requires different types of computation for learning and inference. In contrast, sigprop uses only a single type of computation and is backwardpass unlocked and forwardpass unlocked.

\section{Signal Propagation} \label{sec:ffl}
The premise of signal propagation (sigprop) is to reuse the forward path to map an initial learning signal into targets at each layer for updating parameters. The network is shown in Fig. \ref{fig:fflversions}a; notice that training uses the same forward path as inference, except that instead of only feeding the network the input $x$, we also feed it $c$ the learning signal. The learning signal is some context $c$, e.g. the label in supervised learning. The learning signal and the input can have different shapes, e.g. a supervised label is a single integer and the input is an image. The target generator projects the learning signal $c$ and the first hidden layer projects the input $x$ to both have the same shape (dense signal) or concordant shapes (sparse signal Sec \ref{sec:sparselearning}) to be processed by the network, e.g. the target generator projects the label to have the same shape as the input or even the first hidden layer. After which, the forward pass during training proceeds the same way as inference, except with $x$ and $c$ as the new inputs instead of only the original input $x$.

We provide a framework for any given input $x$ or learning signal $c$, not only for supervised learning with labels. For example, in regression tasks, the inputs $x$ and outputs $y$ commonly have the same type and shape; so, by using the output training targets $y^*$ as the learning signal $c$, the target generator and first hidden layer can be the same (weight sharing). Nonetheless, the focus here is supervised learning.

In the following sub-sections, we start with the general training procedure \ref{sec:sp-train}, then prediction for both training and inference \ref{sec:sp-pred}, the loss for training \ref{sec:sp-loss}, and details of target generators \ref{sec:sp-targen}.

\subsection{Training} \label{sec:sp-train}
The forward pass starts with the input $x$, a learning signal $c$, and the target generator. Assume the network has two hidden layers, as shown in Figures \ref{fig:fflversions}a, where $W_i$ and $b_i$ are weight and bias for layer $i$. Let $S_1$ and $d_1$ be the weight and bias for the target generator. The activation function $f()$ is a non-linearity. Let $(x,y^*)$ be a mini-batch of inputs and labels of $m$ possible classes. We feed $x$ into the first hidden layer to get $h_1$. We create a one-hot vector of each class $c_m$, this is our learning signal, and feed it into the target generator to get $t_1$. Notice that $x$ and $c_m$ have different shapes. Now, $h_1$ and $t_1$ have the same shape.
\begin{align} \label{eq:targen1}
    h_1, t_1 &= f(W_1 x + b_1), f(S_1 c_m + d_{1})\\
    [h_2,t_2] &= f(W_2 [h_1,t_1] + b_2)\\
    [h_3,t_3] &= f(W_3 [h_2,t_2] + b_2)
\end{align}
The outputted $t_1$ is a target for the output of the first hidden layer $h_1$. This target is used to compute the loss $L_1(h_1,t_1)$ for training the first hidden layer and the target generator. Then, the target $t_1$ and the output $h_1$ are fed to the next hidden layer. The forward pass continues this way until the final layer. The final layer and each hidden layer have their own losses:
\begin{align} \label{eq:totalloss}
    J = L(h_1,t_1) + L(h_2,t_2) + L(h_3,t_3)
\end{align}
where $J$ is the total loss for the network. For hidden layers, the loss $L$ can be a supervised loss, such as $L_{pred}$ Eq. \ref{eq:predloss} which is used in Section \ref{sec:exp}. It can also be a Hebbian update rule, such as Eq. \ref{eq:uniupdate} which is used in Section \ref{sec:contime}. For the final layer, the loss $L$ is a supervised loss, such as $L_{pred}$ Eq. \ref{eq:predloss}.

In total, each layer processes its input and input-target to create an output and output-target. The layer compares its output with its output-target to update its parameters. In this way, the layer locally computes its update from a global learning signal. The layer then sends its output and output-target to the next layer which will compute its own update. This processes continues until the final layer has computed its update and produces the network's output (prediction). From this procedure collectively, the network learns to process the input to produce an output, and at the same time, learns to make an initial learning signal into a useful training target at each hidden layer and final layer. In other words, the network itself, which is the forward path, takes on the role of the feedback connectivity in producing a learning signal for each layer. This makes sigprop compatible with models of learning where backward connectivity is limited, such as in the brain and learning in hardware (e.g. neuromorphic chips).

\subsection{Prediction for Training and Inference}  \label{sec:sp-pred}
In training, the prediction $y$ is formed by comparing the final layer's output $h_3$ with its target $t_3$ (Output Target) - Fig \ref{fig:fflversions}a. For inference, the same procedure may be used if group targets, such as class labels, are available. However, no target of any kind is needed for inference - Fig \ref{fig:fflversions}b. Instead, a classification layer may be used with no effect on performance (Classification Layer) Fig \ref{fig:fflversions}b. In general, the last layer may be any type of prediction layer, such as a classification layer or the output layer for regression tasks. With a prediction layer, inference for classification, regression, or any task proceeds as usual, without using a target. We describe both version of sigprop below.

\subsubsection*{\bf Output Target, Fig \ref{fig:fflversions}a} The network's prediction $y$ at the final layer is formed by comparing the output $h_3$ and outputted target $t_3$ (Fig \ref{fig:fflversions}a):
\begin{align} \label{eq:netpred}
    y = y_3 &= O(h_3,t_3)
\end{align}
where $O$ is a comparison function. Two such comparison functions are the dot product and L2 distance. We use the less complex $O_{dot}$,
\begin{align} \label{eq:predcomp}
    O_{dot}(h_i,t_i) &= h_i \cdot t_i^T\\
    O_{l2}(h_i,t_i) &= \sum_k ||t_i[i,1,k] - h_i[1,j,k]||^2_2
\end{align}
but both versions give similar performance using the losses in Section \ref{sec:sp-loss}. Each hidden layer can also output a prediction, these are known as early exits (faster responses from earlier layers during inference):
\begin{align} \label{eq:allpred}
    y = y_i &= O(h_i,t_i)
\end{align}

\subsubsection*{\bf Classification Layer, Fig \ref{fig:fflversions}b}
The final layer of the network may be replaced with the standard output layer used in neural networks, e.g. the classification layer for supervised learning, as shown in Fig \ref{fig:fflversions}b. This simplifies predictions during inference, matching standard neural network design. In this case, the learning signal $c$ (e.g. labels in supervised learning) would be projected to the final layer of the network, as per standard training of networks. The target $t_3$ is no longer used during inference to form $y$, so neither is the context generator.

\subsection{Training Loss} \label{sec:sp-loss}
In sigprop, losses compare neurons with themselves over different inputs and with each other. The $L_{pred}$ is the basic loss we use.

\subsubsection*{\bf Prediction Loss}
The prediction loss is a cross entropy loss using a local prediction, Eq \ref{eq:allpred}. The local prediction is from a dot product between the layer's local targets $t_i$ and the layer's output $h_i$. Given a hidden layer's local targets $t_i = (t_i^1,\dots,t_i^m)$ and a size $n$ mini-batch of outputs $h_i = (h_i^1,\dots,h_i^n)$ of the same hidden layer:
\begin{align} \label{eq:predloss}
    L_{pred}(h_i,t_i) &= \text{CE}(y^*_i, -O_{dot}(h_i,t_i))
\end{align}
where $h_i$ and $t_i$ have the same size output dimension. The cross entropy loss (CE) uses $y^*_i$, which is a reconstruction of the labels $y^*$ at each layer $i$ from the positional encoding of the inputs $x$ and context $c_m$., starting from the activations $h_1$ and targets $t_1$ formed at the first hidden layer. In particular, we form a new batch $[h_1,t_1]$ by interleaving $h_1$ and $t_1$ such that each sample's activations in $h_1$ is concatenated after its corresponding target $t_1$. Then, at each layer $i$, we assign a label for each sample $h_{ij}$ depending on which target $t_{ik}$ the sample came after, where $0 \le k < j$. Many different encodings are available, depending on the task and target generator. An alternative is to use the approach in Section \ref{sec:contime} which merges the context $c$, and therefore generated targets $t_1$, with the inputs $x$ to form a single combined input $xt$, an input-target \ref{sec:sp-tarinp}, and then either compares them with each other or uses an update rule over multiple iterations. The second option is natural for continuous networks where multiple iterations (e.g time steps) can support robust update rules.

\subsection{Target Generators} \label{sec:sp-targen}
The target generator takes in a learning signal as some context $c$ to condition learning on and then produces the initial target, which is fed forward through the network to produce targets at each hidden layer. There are many possible formulations of the target generator, such as: fixed or learned, projecting to input or first hidden layer, and sharing weights with the first hidden layer. We recommend deciding based on the task, selected learning signal(s), and implementation constraints. For example, in segmentation tasks where outputs have the same shape as the inputs, we can use the output training segmentation targets for the learning signal and have the target generator share weights with the first hidden layer. We describe three formulations below to address different learning scenarios, particularly hardware constrained, continuous, and spike-time learning.

\subsubsection*{\bf Target-Only, Fig \ref{fig:fflversions}a,b} This is the version described in Eq. \ref{eq:targen1} and conditions only on the class label. This version of the target generator can interfere with batch normalization statistics as $h_1$ and $t_1$ do not necessarily have similar enough distribution. Batch normalization statistics may be disabled or be put in inference mode when processing the targets, therefore only collecting statistics on the input.

\subsubsection*{\bf Target-Input, Fig \ref{fig:fflversions}a,b} \label{sec:sp-tarinp} Another context we condition on is the class label and input. We feed a one-hot vector of the labels $y^*_m$ through the target generator to produce a scale and shift for the input. We take the scaled and shifted output as the target for the first hidden layer.
\begin{align} \label{eq:targen3}
    t_1 &= h_1 f(S_1 c_m + d_1) + f(S_2 c_m + d_2)
\end{align}
The target $t_1$ is now more closely tied to the distribution of the input. We found that this formulation of the target works better with batch normalization. Even though this version has similar performance to Eq. \ref{eq:targen1}, it increases memory usage as each input will have its own version of the targets.

\begin{figure}[!t]
\centering
\includegraphics[scale=.36,trim = 0 5 0 0,clip]{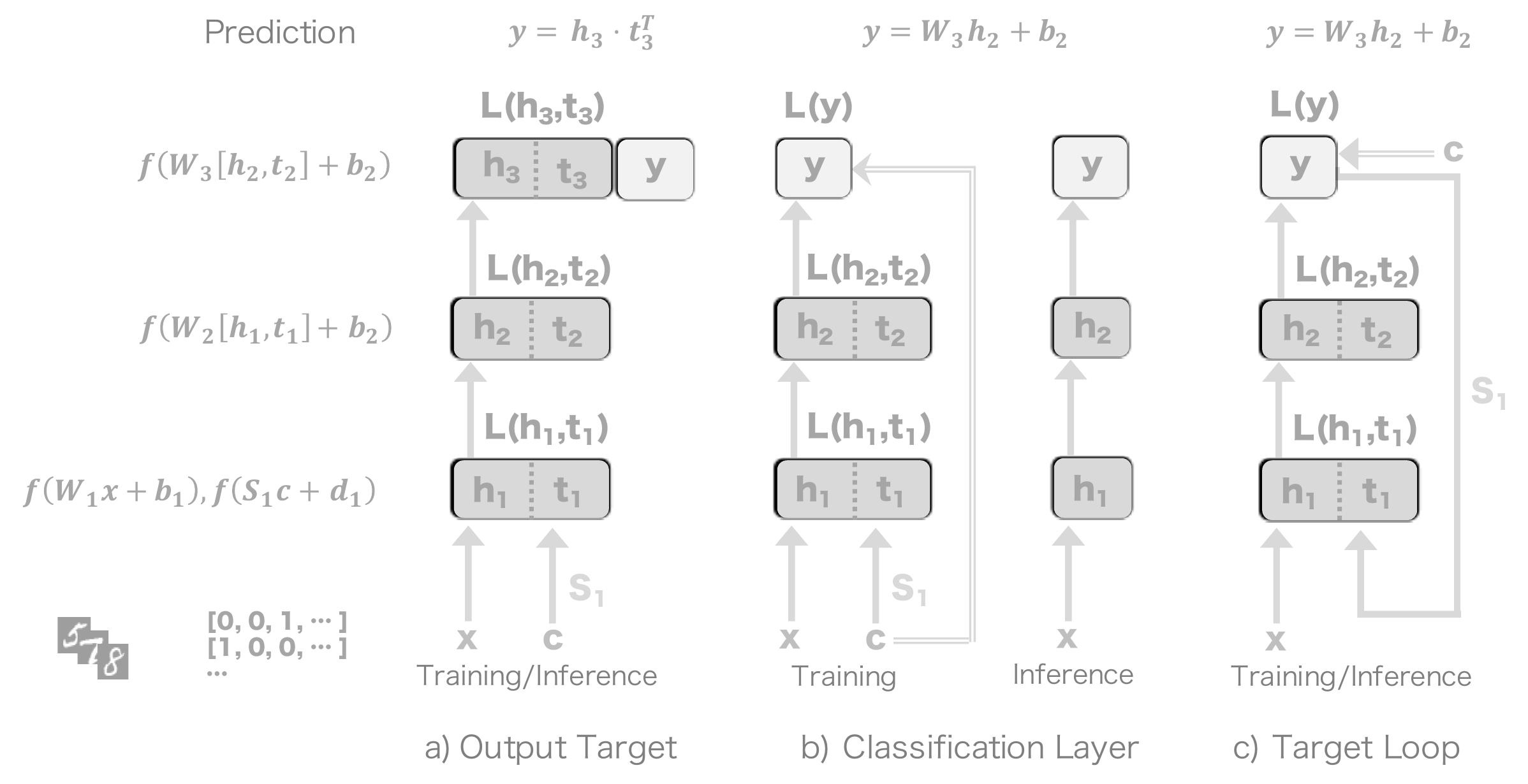}
\caption{Different Versions of sigprop (SP). \textbf{a)} For sigprop, the prediction $y$ is formed by taking $t_3$ with $h_3$. sigprop does not need a classification layer. \textbf{b)} However, a classification layer may be used without effecting performance. In this case, the last hidden layer's outputs are sent to the classification layer. The classification layer has a benefit for inference. During inference, the target $t_3$ is no longer needed to make predictions, so the context $c$ and target generator are not used. \textbf{c)} This is the version of sigprop used in Sections \ref{sec:contime} for the continuous rate model. The classification layer feeds back into the input layer creating a feedback loop, so $y$ is the context $c$: $y = c$. This feedback loop allows the target of hidden layers earlier in the network to incorporate information from hidden layers later in the network without incurring the overhead of reciprocal feedback to every neuron. Continuous networks have multiple iterations which is ideal for this version of sigprop. The other versions of sigprop may also be used.}
\label{fig:fflversions}
\end{figure}

\subsubsection*{\bf Target-Loop, Fig \ref{fig:fflversions}c} \label{sec:target-loop} The last option is to incorporate a form of feedback. The immediate choice is to condition on the activations of the predictions $y_3$ and labels $y^*_m$,
\begin{align} \label{eq:targen4}
    t_1 &= f(S_1 y_3 + S_1 y^*_m + d_1)
\end{align}
or using the final layer's output and error $e_3$ with the target $t_3$ to correct it
\begin{align} \label{eq:targen5}
    t_1 &= f(S_1 (h_3 - \eta e_3) + d_1)\\ \nonumber
    &\triangleq f(S_1 (h_3 - \eta \DD{L}{h_3}) + d_1)
\end{align}
where $\eta$ controls how much error $e_3$ to integrate. We use it in Section \ref{sec:contime} for continuous networks.

\subsection{Sparse Learning} \label{sec:sparselearning}
Sigprop can be a form of sparse learning. We reformulate the target generator to produce a sparse target, which is a sparse learning signal. We make the targets $t_i$ as sparse as possible such that at minimum, they can still be taken with each layer's weights $W_i$, via a convolution or dot-product, and then fed-forward through the network. To make the target sparse, we reduce the output size of $S_i$ in the target generator. We use sparse learning throughout this paper, except when otherwise written.

For convolutional layers, the output size of $S_i$ is made the same size as the weights. For example, let there be an input of $32x28x28$ and a convolutional hidden layer of $32x16x3x3$, where $32$ is the in-channels, $28x28$ is the width and height of the input, $16$ is the out-channels, and $3x3$ is the kernel. The dense target's shape is $32x28x28$. In contrast, the sparse target's shape is reduced to $10x32x3x3$. As a result, even though convolutional layers have weight sharing, there is no weight sharing when convolving with a sparse target.

For fully connected layers, the output size of $S_i$ is made smaller than input size of the weights. For example, let there be an input of $1024$ and a fully connected hidden layer of $1024x512$ features. The dense target's shape would be $1024$. In contrast, the sparse target's shape is $< 1024$. Then, we resize the target to match the layer input size of $1024$ by filling it with zeros. With the sparse target, the layer is no longer fully connected.

\begin{figure}[!t]
\centering
\includegraphics[scale=.25,trim = 0 5 0 0,clip]{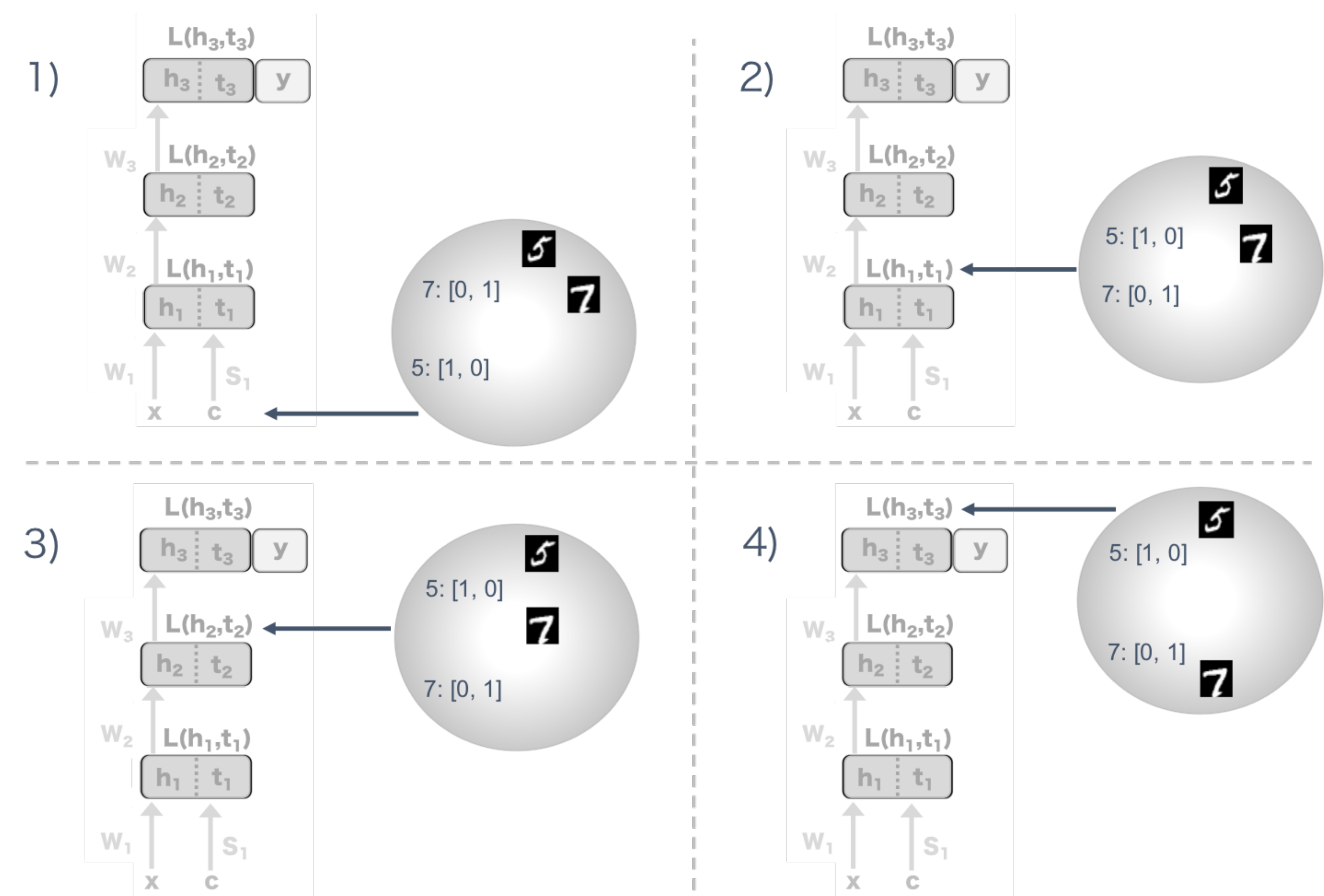}
\caption{Training in sigprop (SP). The learning signals $c$ and inputs $x$ are fed into the network. Then, each layer successively brings the learning signal $5: [1,0]$ closer to the images of $5$, but farther away from learning signal $7: [0,1]$ and images of $7$. The same is done for $7$. Before the first layer \textbf{1)}, the images and learning signal of the same class are not closer to each other than to other classes. In the first layer \textbf{2)}, we nudge $5: [1,0]$ and the image of $5$ closer; the same for $7$. This continues in the following layer \textbf{3)} and then the final layer \textbf{4)}, at which point the learning signal and inputs of the same class are close each other, but farther from the other class. In general, each layer successively bring inputs $x$ and there respective learning signals $c$ closer together than all other inputs and learning signals.}
\label{fig:ffltraining}
\end{figure}

\section{Experiments} \label{sec:exp}
We compare sigprop (SP) with Feedback Alignment (FA) and Local Learning (LL). We also show results for backpropagation (BP) as reference. The models are shown in Figure \ref{fig:models}. FA uses fixed random weights to transport error gradient information back to hidden layers, instead of using symmetric weights. For LL, we show results for two model versions. The first uses BP at the layer level (LL-BP), and the second uses FA in the auxiliary networks to have a backpropagation free model that relaxes learning constraints under backpropagation (LL-FA). LL-FA performs better than using FA or DFA alone. We use LL-BP and LL-FA with predsim losses on the VGG8b architecture \cite{nokland2019training}. We trained several network on the CIFAR-10, CIFAR-100, and SVHN datasets. We used a VGG architecture. The experiments were run using the PyTorch Framework. All training was done on a single GeForce GTX $1080$. For each layer to have a separate loss, the computational graph was detached before each hidden layer to prevent the gradient from propagating backward past the current layer. The target generator was conditioned on the classes, producing a single target for each class.

\textbf{Results for BP, LL-BP, LL-FA, and SP} A batch size of $128$ was used. The training time was $100$ epochs for SVHN, and $400$ epochs for CIFAR-10 and CIFAR-100. ADAM was used for optimization \cite{kingma2014adam}. The learning rate was set to $5e-4$. The learning rate was decayed by a factor of $.25$ at $50\%$, $75\%$, $89\%$, and $94\%$ of the total epochs. The leaky ReLU activation with a negative slope of 0.01 was used \cite{maas2013rectifier}. Batch normalization was applied before each activation function \cite{ioffe2015batch} and dropout after. The dropout rate was $0.1$ for all datasets. The standard data augmentation was composed of random cropping for all datasets and horizontal flipping for CIFAR-10 and CIFAR-100. The results over a single trial for VGG models.

The CIFAR-10 dataset \cite{krizhevsky2009learning} consists of $50000$ $32x32$ RGB images of vehicles and animals with $10$ classes. The CIFAR-100 dataset \cite{krizhevsky2009learning} consists of $50000$ $32x32$ RGB images of vehicles and animals with $100$ classes. The SVHN dataset \cite{netzer2011reading} consists of $32x32$ images of house numbers. We use both the training of $73257$ images and the additional training of $531131$ images.

\subsection{Efficiency}
We measured training time and maximum memory usage on CIFAR-10 for BP, LL-BP, LL-FA, and SP. The version of SP used is \ref{fig:fflversions}b with the $L_{pred}$ loss. The results are summarized in Table \ref{table:cifar10LLEfficiency}. LL and SP training time are measured per layer as they are forwardpass unlocked and layers can be updated in parallel. However, BP is not forwardpass unlocked, so layers are updated sequentially and is therefore necessarily measured at the network level. Measurements are across all seven layers, which is the source of the high variance for LL and SP, and over four hundred epochs of training. To ensure training times are comparable, we compare the epochs at which SP, LL, and BP converge toward their lowest test error. We also include the first epochs that have performance within $0.5\%$ of the best reported performance. All learning algorithms converge within significance of their best performance around the same epoch. Given efficiency per iteration, SP is faster than the other learning algorithms and has lower memory usage.

The largest bottleneck for speed of LL and SP is successive calls to the loss function in each layer. Backpropagation only needs to call the loss function once for the whole network; it optimizes the forward and backward computations for all layers and the batch. SP and LL would benefit from using a larger batch size than backpropagation. The batch size could be increased in proportion to the number of layers in the network. This is only pragmatic in cases where memory can be sacrificed for more speed (e.g. not edge devices).
We also provide per layer measurements in Tables \ref{table:cifar10LLEfficiencyTimeLayers}. At the layer level, SP remains faster and more memory efficient than LL and backpropagation. It is interesting to note that LL and SP tend to be slower and faster in different layers even though both are using the same architecture. For memory, SP uses less memory than LL and BP regardless of the layer. However, there is a general trend for LL and SP: the layers closer to the input have more parameters, so are slower and take up more memory then layers closer to the output.

\begin{table*}[!t]
    \caption{The Training Time Per Sample and Maximum Memory Usage Per Batch Over All Layers for VGG8b}\label{table:cifar10LLEfficiency}
\centering
\begin{tabular}{llllll} \toprule
    &  & \multicolumn{2}{|c|}{Backprop} & \multicolumn{2}{c}{Alternative}  \\
    &  & \multicolumn{1}{|l}{BP} & \multicolumn{1}{l|}{LL-BP} & LL-FA & SP \\ \midrule
    Time (s) & CIFAR-10 & $12.29\pm0.02$ & $8.11\pm14.40$ & $8.50\pm29.86$ & $\boldsymbol{5.91}\pm7.40$ \\
        & CIFAR-100 & $15.34 \pm 1.45$ & $10.20 \pm 28.98$ & $9.44 \pm 28.63$ & $\boldsymbol{6.25} \pm 7.33$ \\
             & SVHN & $148.70 \pm 2.23$ & $95.51 \pm 3617.90$ & $89.32 \pm 1767.26$ & $\boldsymbol{69.74} \pm 1048.54$ \\ \midrule
             Mem (MiB) & CIFAR-10 & $22.00\pm0.00$  & $8.85\pm8.06$ & $13.03\pm10.61$ & $\boldsymbol{6.19}\pm1.57$ \\
             & CIFAR-100  & $27.16 \pm 0.38$  & $11.45 \pm 106.02$ & $5.51 \pm 23.17$ & $\boldsymbol{5.19} \pm 16.72$ \\
              & SVHN & $28.04 \pm 2.68$  & $11.41 \pm 106.03$ & $5.43 \pm 23.04$ & $\boldsymbol{4.91} \pm 16.54$ \\ \midrule
              Best Epoch & CIFAR-10  & $319 (198)$  & $266 (164)$ & $309 (201)$ & $313 (207)$ \\
              & CIFAR-100  & $350 (306)$  & $380 (209)$ & $339 (264)$ & $329 (219)$ \\
              & SVHN  & $98 (11)$  & $41 (7)$ & $93 (23)$ & $88 (34)$ \\ \midrule
              Test Error (\%) & CIFAR-10 & $5.99$  & $\boldsymbol{5.58}$ & $9.02$ & \underline{$8.34$} \\
              & CIFAR-100 & $\boldsymbol{26.20}$ & $29.31$ & $38.41$ & \underline{$34.30$} \\
              & SVHN & $2.19$ & $\boldsymbol{1.77}$ & $2.55$  & \underline{$2.15$} \\ \bottomrule
\end{tabular}
\end{table*}

\begin{table}[!t]
    \caption{The Training Time Per Sample and Maximum Memory Usage Per Batch per Layer on CIFAR-10 for VGG8b}\label{table:cifar10LLEfficiencyTimeLayers}
\centering
\begin{adjustbox}{max width=\columnwidth}
\begin{tabular}{llll} \toprule
     & \multicolumn{1}{|c|}{Backprop} & \multicolumn{2}{c}{Alternative}  \\
     Layer & \multicolumn{1}{|l|}{LL-BP} & LL-FA & SP \\ \midrule
    \multicolumn{4}{c}{Time (s)}  \\
    1 & $7.16 \pm 0.04 $ & $6.21 \pm 0.03 $ & $\boldsymbol{4.48} \pm 0.05 $ \\
        2 & $15.80 \pm 0.07$ & $15.15 \pm 0.09$ & $\boldsymbol{8.95} \pm 0.15 $ \\
        3 & $9.27 \pm 0.04 $ & $\boldsymbol{7.09} \pm 0.02 $ & $10.13 \pm 0.14$ \\
        4 & $9.25 \pm 0.30 $ & $18.40 \pm 0.06$ & $\boldsymbol{7.27} \pm 0.25 $ \\
        5 & $4.93 \pm 0.01 $ & $5.66 \pm 0.04 $ & $\boldsymbol{4.71} \pm 0.05 $ \\
        6 & $7.46 \pm 0.01 $ & $3.93 \pm 0.02 $ & $\boldsymbol{3.44} \pm 0.02 $ \\
        7 & $2.90 \pm 0.00 $ & $3.00 \pm 0.00 $ & $\boldsymbol{2.36} \pm 0.03 $ \\ \midrule
    \multicolumn{4}{c}{Mem (MiB)}  \\
    1,6,7 & $6.12 $ & $10.98$ & $\boldsymbol{5.67}$\\
    2 & $14.50$ & $18.18$ & $\boldsymbol{9.26}$\\
    3 & $9.70 $ & $18.18$ & $\boldsymbol{5.67}$\\
    4,5 & $9.70 $ & $10.97$ & $\boldsymbol{5.67}$\\
    \bottomrule
\end{tabular}
\end{adjustbox}
\end{table}

\subsection{Sparse Local Targets} \label{sec:sparsetargets}
We demonstrate that sigprop (SP) can train train a network with a sparse learning signal. We use the larger VGG8b(2x) architecture to leave more room for possible improvement when using this sparse target. The version of sigprop is \ref{fig:fflversions}b with the $L_{pred}$ loss. We use the CIFAR10 dataset with the same configuration as in Section \ref{sec:exp}. We see that the network's training speed increased and memory usage decreased Fig. \ref{table:cifar10SPEfficiency},\ref{table:cifar10SPEfficiencyLayers}, with negligible change in accuracy.

\begin{table}[!t]
    \caption{Efficiency of Targets Over All Layers on CIFAR-10 for VGG8b(2x). Training Time Per Sample, Maximum Memory Usage Per Batch}\label{table:cifar10SPEfficiency}
\centering
\begin{adjustbox}{max width=\columnwidth}
\begin{tabular}{llllll} \toprule
              &            Dense & Sparse \\ \midrule
              Time (s)             &  $14.48 \pm 54.29$   & $\boldsymbol{9.56} \pm 29.02$ \\
        Mem (MiB)            & $14.04 \pm 6.39$ & $\boldsymbol{10.74} \pm 65.10$ \\
        Best Epoch      & $273 (207)$ & $340 (219)$ \\
              Test Error (\%) &           $7.60$ & $7.71$ \\ \bottomrule
\end{tabular}
\end{adjustbox}
\end{table}

\begin{table}[!t]
    \caption{Efficiency of Targets per Layer on CIFAR-10 for VGG8b(2x). Training Time Per Sample and Maximum Memory Usage Per Batch}\label{table:cifar10SPEfficiencyLayers}
\centering
\begin{adjustbox}{max width=\columnwidth}
\begin{tabular}{lllll} \toprule
    Layer & \multicolumn{4}{c}{Time s (Mem MiB)}  \\
    & \multicolumn{2}{c}{Dense} & \multicolumn{2}{c}{Sparse} \\ \midrule
    1 & $12.85 \pm 5.66 $ &($12.99$) & $\boldsymbol{7.42 }\pm 0.79 $ &($\boldsymbol{6.34} $)  \\
    2 & $21.51 \pm 9.31 $ &($\boldsymbol{20.23}$) & $\boldsymbol{19.70} \pm 0.18$ &($27.53$)  \\
    3 & $18.81 \pm 5.50 $ &($13.02$) & $\boldsymbol{9.30 }\pm 0.39 $ &($\boldsymbol{9.41} $)  \\
    4 & $25.30 \pm 12.97$ &($\boldsymbol{13.02}$) & $\boldsymbol{14.19} \pm 0.12$ &($15.99$)  \\
    5 & $9.69 \pm 1.86 $  &($13.02$) & $\boldsymbol{8.84 }\pm 0.11 $ &($\boldsymbol{9.10} $) \\
    6 & $8.11 \pm 3.16 $  &($13.02$) & $\boldsymbol{5.24 }\pm 0.08 $ &($\boldsymbol{6.15} $) \\
    7 & $5.06 \pm 1.61 $  &($12.99$) & $\boldsymbol{2.25 }\pm 0.07 $ &($\boldsymbol{0.68} $) \\ \bottomrule
\end{tabular}
\end{adjustbox}
\end{table}

\section{In Continuous Time} \label{sec:contime}
We demonstrate that sigprop can train a neural model in the continuous setting using a Hebbian update mechanism, in addition to the discrete setting. Biological neural networks work in continuous time, have no indication of different dynamics in inference and learning, and use Hebbian based learning. Sigprop improves learning in this scenerio by bringing a global learning signal into Hebbian based learning, without the comprehensive feedback connectivity to neurons and layers that previous approaches require, not observed in biological networks. In addition, sigprop improves compatibility for learning in hardware, such as neuromorphic chips, which have resource and design constraints that limit backward connectivity.

In the model presented in this section, the target generator is conditioned on the activations of the output layer to produce a feedback loop - Fig. \ref{fig:fflversions}c. The feedback loop is always active, during training and inference. With this feedback loop, we demonstrate in section \ref{sec:contimemodel} that sigprop provides useful learning signals by bringing forward and feedback loop weights into alignment. In Section \ref{sec:contimelearn}, we measured the performance of this model on the MNIST and Fashion-MNIST datasets \cite{lecun1998mnist,xiao2017fashion}.

\subsection{A Continuous Recurrent Neural Network Model} \label{sec:contimemodel}
The learning framework, Equilibrium Propagation (EP), proposed in \cite{scellier2017equilibrium} is one way to introduce physical time in deep continuous learning and have the same dynamics in inference and learning, avoiding the need for different hardware for each. EP has been used with symmetric or random feedback weights. We combine Sigprop with EP such that there are no additional constraints on learning, beyond the Hebbian update. We trained deep recurrent networks with a neuron model based on the continuous Hopfield model \cite{hopfield1984neurons}:
\begin{align} \label{eq:neuronmodel}
    \DDT{s_j} =& \DD{\rho(s_j)}{s_j} (\sum_{i \to j} w_{ij} \rho(s_i) + \sum_{i \in O \to j \in I} w_{ij} \rho(s_i) + b_j)\\ \nonumber
            &- \frac{s_j}{r_j} - \beta \sum_{j \in O} {(s_j - d_j)}
\end{align}
where $s_j$ is the state of neuron $j$, $\rho(s_j)$ is a non-linear monotone increasing function of it's firing rate, $b_j$ is the bias, $\beta$ limits magnitude and direction of the feedback, $O$ is the subset of output neurons, $I$ is the subset of input receiving neurons, and $d_j$ is the target for output neuron $j$. The input receiving neurons, $s_j \in I$, are the neurons with forward connections from the input layer. The networks are entirely feedforward except for the final feedback loop from the output neurons $s_i \in O$ to the input receiving neurons $s_j \in I$. All weights and biases are trained. The weights in the feedback loop connections may be fixed or trained. The output neurons receive the $L_2$ error as an additional input which nudges the firing rate towards the target firing rate $d_j$. The target firing rate $d_j$ is the one-hot vector of the target value; all tasks in this section are classification tasks.

The EP learning algorithm can be broken into the free phase, the clamped phase, and the update rule. In the free phase, the input neurons are fixed to a given value and the network is relaxed to an energy minimum to produce a prediction. In the clamped phase, the input neurons remain fixed and the rate of output neurons $s_j \in O$ are perturbed toward the target value $d_j$, given the prediction $s_j$, which propagates to connected hidden layers. The update rule is a simple contrastive Hebbian (CHL) plasticity mechanism that subtracts $s_i^0 s_j^0$ at the energy minimum (fixed point) in the free phase from $s_i^\beta s_j^\beta$ after the perturbation of the output, when $\beta > 0$: 
\begin{align} \label{eq:uniupdate}
    \Delta W_{ij} \propto \rho(s_i) \DD{}{\beta} (\rho(s_j)) \approx \frac{1}{\beta} \rho(s_i^0) (\rho(s_j^\beta) - \rho(s_j^0))
\end{align} 
The clamping factor $\beta$ allows the network to be sensitive to internal perturbations. As $\beta \to +\infty$, the fully clamped state in general CHL algorithms is reached where perturbations from the objective function tend to overrun the dynamics and continue backwards through the network.

\subsection{Signal Propagation Provides Useful Learning Signals} \label{sec:contimelearn}
We look at the behavior of our model during training and how the feedback loop drives weight changes. Precise symmetric connectivity was thought to be crucial for effective error delivery \cite{rumelhart1986learning}. Feedback Alignment, however, showed that approximate symmetry with reciprocal connectivity is sufficient for learning \cite{guerguiev2017towards,liao2016important,lillicrap2016random}. Direct Feedback Alignment showed that approximate symmetry with direct reciprocal connectivity is sufficient. In the previous sections, we showed that no feedback connectivity is necessary for learning. Here, we conduct an experiment to show that the same approximate symmetry is found in sigprop.

\begin{figure*}[!t]
    \begin{center}
    {
        \setlength{\tabcolsep}{0pt}
        \renewcommand{\arraystretch}{0}
        \scriptsize
    \begin{tabular}{rcccc}
    \begin{turn}{90}
        \hspace{3.55cm} \textbf{Learned}
    \end{turn} &&
    \includegraphics[scale=.12,trim = 0 0 0 0,clip]{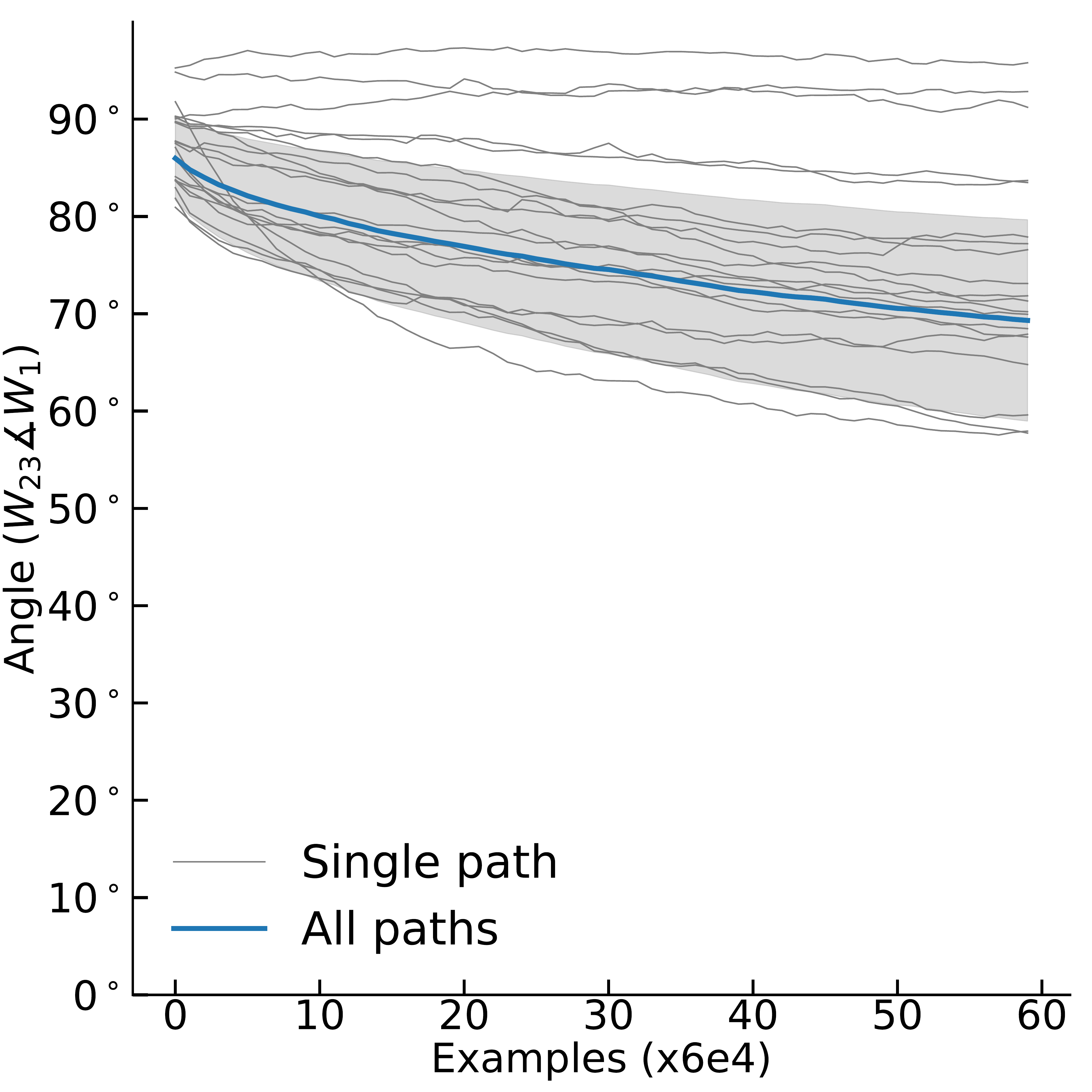} &
    \includegraphics[scale=.12,trim = 0 0 0 0,clip]{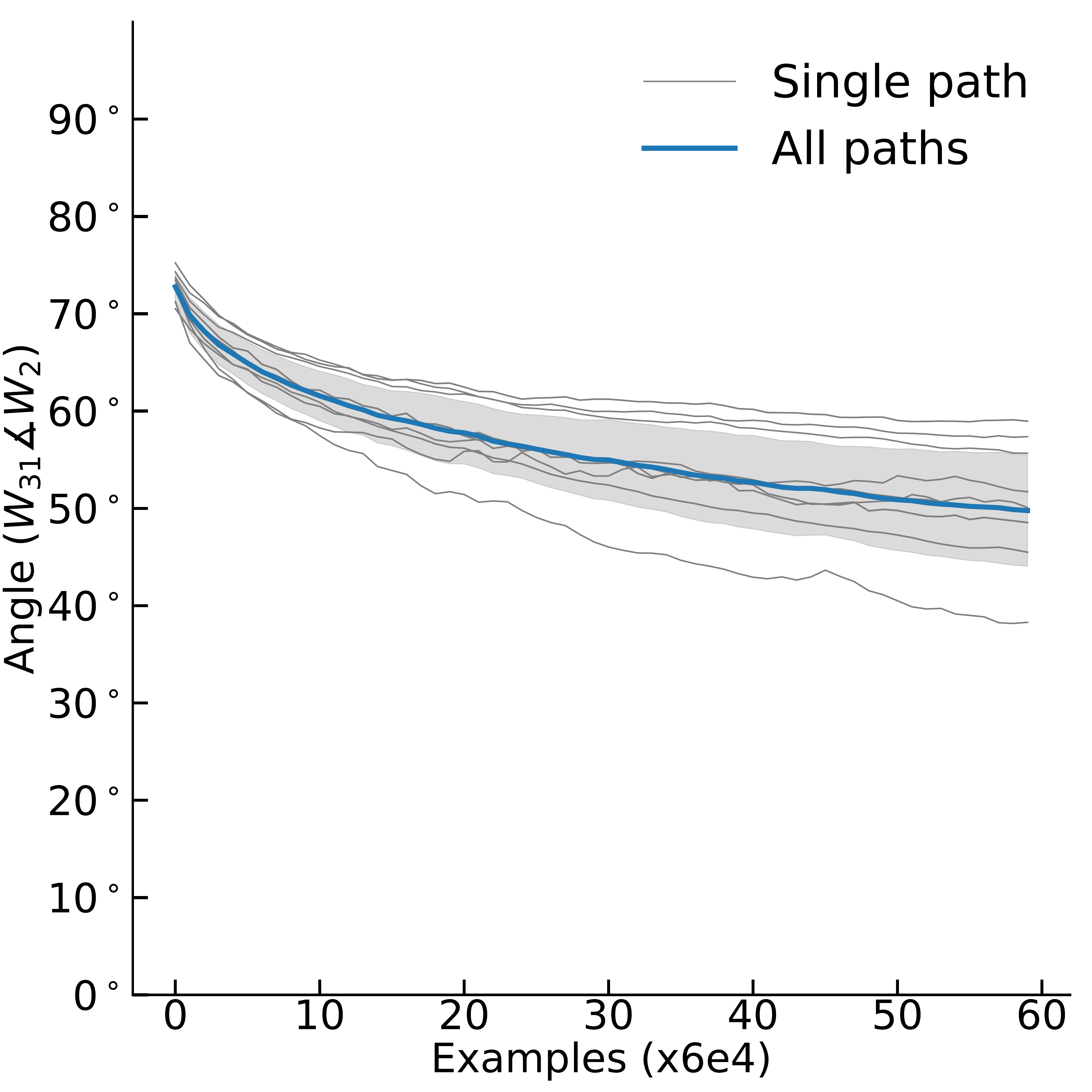} &
    \includegraphics[scale=.12,trim = 0 0 0 0,clip]{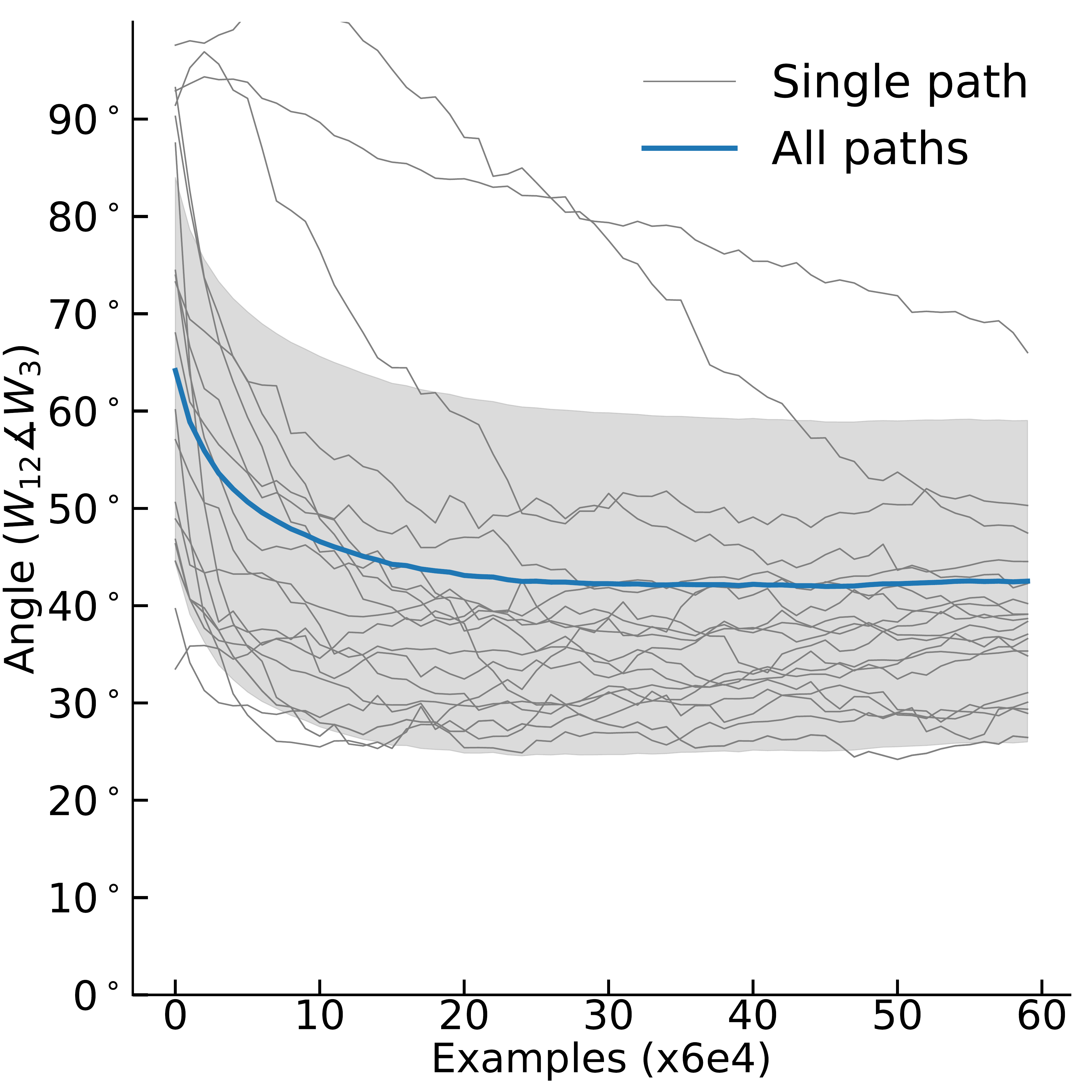} \\
    \vspace{.1cm}&&&&\\
    &&
    \textbf{$\measuredangle$ 1st Hidden Layer} &
    \textbf{$\measuredangle$ 2nd Hidden Layer} &
    \textbf{$\measuredangle$ Loop Back Layer}
    \end{tabular}
    }
    \end{center}
\caption{Signal Propagation updates bring weights into alignment within $90^\circ$, approaching backpropagation symmetric weight alignment. Sigprop provide useful targets for learning. The weight alignment for a network with two hidden layers $W_1$ and $W_2$ and one loop back layer $W_3$ is shown. The weight matrices form a loop in the network and come into alignment with each other during training on the Fashion-MNIST dataset. Each weight matrix aligns with the product of the other two weights forming the network loop. $W_{xy} \measuredangle W_z$ means the angle between weight $z$ and the matrix multiplication of the weights $x$ and $y$. \textbf{Learned}) The loop back layer is trained. However, even a fixed loop back layer reaches a similar angle of alignment. \textbf{Layers}) The loop back layer converges before the 1st and 2nd hidden layers can. The 1st hidden layer is the least aligned with the 2nd hidden layer and the loop back layer because it is dominated by the input signal. The alignment angles are taken for every sample and error bars are one standard deviation.}
    \label{fig:angleloop}
    \vspace{-2.5mm}
\end{figure*}

We provide evidence that sigprop brings weights into alignment within $90^\circ$, known as approximate symmetry. In comparison, backpropagation has complete alignment between weights, known as symmetric connectivity. Note that this is not a measure of approximation to backpropagation - sigprop is a new and different approach; instead, this is a measure of the quality of the learning signal in deeper layers, contextualized by observations of learning with backpropagation, particularly symmetry. In this experiment, the sigprop network architecture forms a loop, so all the weights serve as both feedback and feedforward weights. For a given weight matrix, the feedback weights are all the weights on the path from the downstream error to the presynaptic neuron. In general, this is all the other weights in the network loop. The weight matrices in the loop evolve to align with each other as seen in Fig. \ref{fig:angleloop}. More precisely, each weight matrix roughly aligns with the product of all the other weights in the network loop. In Fig. \ref{fig:angleloop}, the weight alignment for a network with two hidden layers $W_1$ and $W_2$ and one loop back layer $W_3$ is shown.

Information about $W_3$ and $W_1$ flows into $W_2$ as roughly $W_3 W_1$, which nudges $W_2$ into alignment with the rest of the weights in the loop. From equation \ref{eq:uniupdate}, \({W_2 \propto \rho(\vec{s}_2^0) (\rho(\vec{s}_3^\beta) - \rho(\vec{s}_3^0))}\) where \({\vec{s}_2 \gets \rho(\vec{s}_1) W_1}\), which means information about $W_1$ accumulates in $W_2$. Similarly, \({W_1 \propto \rho(\vec{s}_1^0) (\rho(\vec{s}_2^\beta) - \rho(\vec{s}_2^0))}\), except since the network architecture is a feedforward loop, \({\vec{s}_1 \gets \rho(\vec{s}_3) W_3}\), which means information about $W_3$ accumulates in $W_1$. The result is shown in column c of the bottom row of Fig. \ref{fig:angleloop}, where a weight matrix is fixed and the rest of the network's weights come into alignment with the fixed weight. Notice that $W_3 W_1$ has the same shape as $W_2^{T}$ and serves as it's `feedback' weight.

\subsection{Classification Results}
We provide evidence that sigprop with EP has comparable performance to EP with symmetric weights, and report the performance results of the experiment in the previous section. A two and another three layer architecture of $1500$ neurons per layer were trained. The two layer architecture was run for sixty epochs and the three layer for one hundred and fifty epochs. The best model during the entire run was kept. On the MNIST dataset \cite{lecun1998mnist}, the generalization error is $1.85-1.90\%$ for both the two layer and three layer architectures, an improvement over EP's $2-3\%$. The best validation error is $1.76-1.80\%$ and the training error decreases to $0.00\%$. To demonstrate that sigprop provides useful learning signals in the previous section, we trained the network on the more difficult Fashion-MNIST dataset \cite{xiao2017fashion}. The generalization error is $11.00\%$. The best validation error is $10.95\%$ and the training error decreases to $2\%$.

\section{Spiking Neural Networks} \label{sec:spike}
We demonstrate that sigprop can train a spiking neural model with only the voltage (spike), and improves the hardware compatibility of surrogate functions by reducing them to local update rules. This is an improvement over backpropagation based approaches as they: struggle to learn with only the voltage; require going backward through non-derivable, non-continuous spiking equations; and require comprehensive feedback connectivity - all of which are problematic for hardware and biological models of learning \cite{eshraghian2021training,kheradpisheh2020temporal,bouvier2019spiking}.

Spiking is the form of neuronal communication in biological and hardware neural networks. Spiking neural networks (SNN) are known to be efficient by parallelizing computation and memory, overcoming the memory bottleneck of Artificial Neural Networks (ANN) \cite{backus1978can,horowitz20141,mahapatra1999processor}. However, SNNs are are difficult to train. A key reason is that spiking equations are non-derivable, non-continuous and spikes do not necessarily represent the internal parameters, such as membrane voltage of the neuron before and after spiking \cite{bouvier2019spiking}. Spiking also has multiple possible encodings for communication when considering time which are non-trivial, whereas artificial neural networks (ANN) have a single rate value for communication \cite{bouvier2019spiking}. One approach to training SNNs is to convert an ANN into a spiking neural network after training \cite{cao2015spiking,diehl2015fast,rueckauer2017conversion}. Another approach is to have an SNN in the forward path, but have a backpropagation friendly surrogate model in the backward path, usually approximately making the spiking differentiable in the backward path to update the parameters \cite{bouvier2019spiking,mohemmed2012span,yin2017algorithm}.

We trained SNNs with sigprop. The target is forwarded through the network with the input, so learning is done before the spiking equation. That is, we do not need to differentiate a non-derivable, non-continuous spiking equation to learn. Also, the SNN has the same dynamics in inference and learning and has no reciprocal feedback connectivity. This makes sigprop ideal for on-chip, as well as off-chip, training of spiking neural networks. We measure the performance of this model on the MNIST and Fashion-MNIST datasets.

\subsection{Spiking Neural Network}

We train a convolutional spiking neural network with Integrate-and-Fire (IF) nodes, which are treated as activation functions. The IF neuron can be viewed as an ideal integrator where the voltage does not decay. The subthreshold neural dynamics are:
\begin{align} \label{eq:IFnmodel}
    v_i^t = v_i^{t-1} + h_i^t
\end{align}
where  $v_i^t$ is the voltage at time $t$ for neurons of layer $i$ and $h_i^t$ is the layer's activations. The surrogate spiking function for the IF neuron is the arc tangent
\begin{align} \label{eq:IFnsur}
    g(x) = \frac{1}{\pi} \arctan(\pi x) + \frac{1}{2}
\end{align}
where the gradient is defined by
\begin{align} \label{eq:IFnsurdx}
g'(x) = \frac{1}{1 + (\pi x)^2}
\end{align}
The neuron spikes when the subthreshold dynamics reach $0.5$ for sigprop, and $1.0$ for BP and Shallow models. All models is simulated for $4$ time-steps, directly using the subthreshold dynamics. The SNN has $4$ layers. The first two are convolutional layers, each followed by batch normalization, an If node, and a $2x2$ maxpooling. The last two layers are fully connected, with one being the classification layer. The output of the classification layer is averaged across all four time steps and used as the network output. ADAM was used for optimization \cite{kingma2014adam}. The learning rate was set to $5e-4$. Cosine Annealing \cite{loshchilov2016sgdr} was used as the learning rate schedule with the maximum number of iterations $T_{max}$ set to $64$. The models are trained on the MNIST and Fashion-MNIST datasets for $64$ epochs using a batchsize of $128$. We use automatic mixed precision for $16$-bit floating operations, instead of the only the full $32$-bit. The reduced precision is better representative of hardware limitations for learning. We use the classification layer version of sigprop Fig. \ref{fig:fflversions}b.

\subsection{Results}
We compare four spiking models on the MNIST and Fashion-MNIST datasets - Table. \ref{table:spikingfmnist}. The BP model propagates backward through the spiking equations at each layer using a differentiable surrogate. The Shallow model only trains the classification layer. The SP Surrogate model uses the same differentiable surrogate as BP does, but SP propagates forward through the network and therefore does not need to go through the spiking equation to deliver a learning signal. That is, the parameter update and surrogate are before or perpendicular to spiking, possibly as separate compartment. Finally, the SP voltage model uses the neuron's voltage (i.e. directly uses the spiking equation) to calculate the loss and update the parameters, no surrogate is used.

In contrast, BP based learning (without considerable modifications and additions) struggles when only using the voltage for learning \cite{eshraghian2021training,kheradpisheh2020temporal}. A differentiable nonlinear function estimating the spiking behavior (i.e. surrogate) is necessary for reasonable performance in BP learning. A surrogate is also necessary for sigprop to come close to BP surrogate performance. Even without a surrogate, the SP Voltage model is able to train the network significantly better than the Shallow model. To the best of our knowledge, sigprop is the only learning framework with a global supervised (unsupervised, reinforcement) learning signal that satisfies requirements for hardware (on-chip) learning \cite{bouvier2019spiking,davies2021advancing}.

\begin{table}[!t]
    \caption{The Test Error for a Spiking Convolutional Neural Network.}\label{table:spikingfmnist}
\centering
\begin{adjustbox}{width=\columnwidth}
\begin{tabular}{lllllll} \toprule
    & \multicolumn{1}{c|}{BP} & & \multicolumn{2}{|c}{SP} \\
    & \multicolumn{1}{c|}{Surrogate} & Shallow & \multicolumn{1}{|c}{Surrogate} & Voltage \\ \midrule
    Fashion-MNIST & $6.70$ & $16.42$ & $9.51$ & $10.68$ \\
    MNIST        & $0.84$ & $7.24$ & $1.01$ & $2.63$ \\ \bottomrule
\end{tabular}
\end{adjustbox}
\end{table}

\section{Discussion and Conclusion}
Alternative learning algorithms to backpropagation relax constraints on learning under backpropagation, such as feedback connectivity, weight transport, multiple types of computations, or a backward pass. This is done to improve training efficiency, lowering time or memory, or enabling distributed or parallel execution; and, to improve compatibility with biological and hardware learning models. However, relaxing constraints negatively impacts performance. So, alternatives try varying relaxations or supplementary modifications and additions in an attempt to retain the performance found under backpropagation. For instance, the best performing and least constrained alternative algorithm, LL-FA, uses a layer-wise loss and random feedback to relax constraints, but adds layer-wise auxiliary networks to retain performance. In contrast, sigprop has no constraints on learning, beyond the inference model, and without constraining (e.g. layer-wise) additions or modifications.

We demonstrated that sigprop has faster training times and lower memory usage than BP, LL-BP, and LL-FA. The reason sigprop is more efficient than BP is clear, sigprop is forwardpass unlocked while BP is backwardpass locked. For LL-BP and LL-FA, sigprop is more efficient as it has fewer layers for learning, it has no auxiliary networks. LL-BP has 2 auxiliary layers for every hidden layer. LL-FA has 3 auxiliary layers for every hidden layer. 
In Section \ref{sec:sparsetargets}, we showed that sparse targets, which have a much smaller size than the hidden layer outputs, are able to train the hidden layer as well as dense targets, which have the same size as the hidden layer outputs. A key feature of learning in the brain and biological neural networks is sparsity. A small fraction of the neurons weigh in on computations and decision making. It is encouraging that sigprop is able to learn just as well with a sparse learning signal.

In Section \ref{sec:contime}, we applied sigprop to a time continuous model using a Hebbian plasticity mechanism to update weights, demonstrating sigprop has dynamical and structural compatibility with biological and hardware learning. With this continuous model, we also showed that sigprop is able to provide useful learning signals. While sigprop improves the performance of EP, the Fashion-MNIST results demonstrate that there is room for growth. One problem may be that the layers on the path from the input to the output have their weight updates dominated by the input, so are struggling to come into alignment with the loopback layer. In future work, we will compensate to increase alignment.

In Section \ref{sec:spike}, we demonstrated a key feature of sigprop not seen in other global learning algorithms: sigprop does not need to go through a non-derivable, non-continuous spiking equation to provide a learning signal to hidden layers. This makes sigprop ideal for hardware (on-chip) learning. Furthermore, sigprop is able to train an SNN using spikes (voltage), which backpropagation struggles to do, and at a reduced $16$-bit precision. So, no additional complex circuitry is necessary. This makes on-chip global learning (e.g supervised or reinforcement) more plausible with sigprop, whereas the complex neuron and synaptic models of previous supervised learning algorithms are impractical \cite{bouvier2019spiking,davies2021advancing}. This is in addition to sigprop not having architectural requirements for learning and having the same type of computation for learning and inference, which on their own address hardware constraints restricting the use of previous supervised learning algorithms \cite{bouvier2019spiking,davies2021advancing}. We are working to implement sigprop on hardware neural networks.

We demonstrated signal propagation, a new learning framework for propagating a learning signal and updating neural network parameters via a forward pass. Our work shows that learning signals can be fed through the forward path to train neurons. In biology, this means that neurons who do not have feedback connections can still receive a global learning signal through their incoming connections. In hardware, this means that global learning (e.g supervised or reinforcement) is possible even though there is no backward connectivity. At its core, sigprop re-uses the forward path to propagate a learning signal and generate targets. With this combination, there are no structural or computational requirements for learning, beyond the inference model. Furthermore, the network parameters are updated as soon as they are reached by a forward pass. So, sigprop learning is ideal for parallel training of layers or modules. In total, we presented learning models across a spectrum of learning constraints, with backpropagation being the most constrained and signal propagation being the least constrained. Signal propagation has better efficiency, compatibility, and performance than more constrained learning algorithms not using backpropagation.

{
\bibliographystyle{IEEEtran}
\bibliography{feedforwardenergy}
}

\clearpage
{\appendix[Additional Results]

\begin{table}[h]
\caption{The test error for BP, FA, DFA, and SP (\textbf{Best} vs \underline{BP})}
    \label{table:perfFA}
\centering
\begin{adjustbox}{width=\columnwidth}
\begin{tabular}{lllllll} \toprule
    Dataset & Network & & BP & FA & DFA & SP \\ \midrule

MNIST & FC &  2x800      & $\underline{1.60}\pm0.06$  & $\boldsymbol{1.64}\pm0.03$  & $1.74\pm0.08$ & $1.71\pm0.03$\\
    && 3x800      & $1.75\pm0.05$  & $\boldsymbol{1.66}\pm0.09$  & $1.70\pm0.04$ & $1.70\pm0.04$\\
    && 4x800      & $1.92\pm0.11$  & $\boldsymbol{1.70}\pm0.04$  & $1.83\pm0.07$ & $\boldsymbol{1.70}\pm0.04$\\
    && 2x800  DO  & $\underline{1.26}\pm0.03$  & $1.53\pm0.03$  & $1.45\pm0.07$ & $\boldsymbol{1.38}\pm0.03$\\  \midrule

    CIFAR-10 & FC & 3x1000 DO  & $\underline{42.20} \pm 0.2$ & $46.90 \pm 0.3$ & $42.90 \pm 0.2$  & $\boldsymbol{42.62}\pm0.16$\\

    & CONV &        & $\underline{22.50} \pm 0.4$ & $27.10 \pm 0.8$ & $26.90 \pm 0.5$ & $\boldsymbol{24.75}\pm0.40$ \\ \midrule

    CIFAR-100 & FC & 3x1000 DO  & $\underline{69.80}\pm0.1$   & $75.30\pm0.2$   & $73.10\pm0.1$  & $\boldsymbol{70.30}\pm0.19$\\

    & CONV &         & $\underline{51.70}\pm0.2$ & $60.50\pm0.3$ & $59.00\pm0.3$ & $\boldsymbol{57.01}\pm0.42$\\
    \bottomrule
\end{tabular}
\end{adjustbox}
\end{table}

    We trained several networks using BP, FA, DFA, and SP on the MNIST, CIFAR-10, and and CIFAR-100. We used fully-connected architectures (FC), and a small convolutional architecture (CONV) architecture. Note, feedback alignment based algorithms (FA and DFA) do not scale well; they are combined them with LL, or another learning model, to achieve reasonable performance.

\vfill

\end{document}